\documentclass{article}
\usepackage[preprint]{neurips_2025}
\usepackage{tabularx}
\usepackage{microtype}
\usepackage{hyperref}
\usepackage{url}
\usepackage{booktabs}
\usepackage{graphicx}
\usepackage{lineno}
\usepackage{multirow}
\usepackage{makecell}
\usepackage[table]{xcolor}
\usepackage{enumitem}
\usepackage{wrapfig}
\usepackage{bbm}
\usepackage{dsfont}
\usepackage{amsmath}
\usepackage[most]{tcolorbox}  

\usepackage{color-edits}
\addauthor{zw}{orange}
\addauthor{sx}{green}

\usepackage{xspace}
\newcommand{\tm}{\textsc{ToolMem}\xspace}
\newcommand{\gr}{\textsc{Generic}\xspace}
\newcommand{\fs}{\textsc{Few-Shot}\xspace}
\newcommand{\gb}{\textsc{GenAI-Bench}\xspace}
\newcommand{\bb}{\textsc{BiGGen Bench}\xspace}

\definecolor{darkblue}{rgb}{0, 0, 0.5}
\hypersetup{colorlinks=true, citecolor=darkblue, linkcolor=darkblue, urlcolor=darkblue}

\title{\tm: Enhancing Multimodal Agents with \\Learnable Tool Capability Memory}

\author{%
  Yunzhong Xiao$^{1}$ \quad
  Yangmin Li$^{1}$ \quad
  Hewei Wang$^{1}$ \quad
  Yunlong Tang$^{2}$ \quad
  Zora Zhiruo Wang$^{1}$ \\
  $^{1}$Carnegie Mellon University \quad
  $^{2}$University of Rochester \\
  \texttt{yunzhonx@alumni.cmu.edu, yangmin2@alumni.cmu.edu, heweiw@alumni.cmu.edu,} \\
  \texttt{yunlong.tang@rochester.edu, zhiruow@andrew.cmu.edu} \\
}

\begin{document}

\maketitle

\begin{abstract}
Agents utilizing tools powered by large language models (LLMs) or vision-language models (VLMs) have demonstrated remarkable progress in diverse tasks across text and visual modalities.
Unlike traditional tools such as calculators, which give deterministic outputs, neural tools perform uncertainly across task scenarios. While different tools for a task may excel in varied scenarios, existing agents typically rely on fixed tools, thus limiting the flexibility in selecting the most suitable tool for specific tasks. 
In contrast, humans snowball their understanding of the capabilities of different tools by interacting with them, and apply this knowledge to select the optimal tool when solving a future task.
To build agents that similarly benefit from this process, we propose \tm that enables agents to develop memories of tool capabilities from previous interactions, by summarizing their strengths and weaknesses and storing them in memory; at inference, the agent can retrieve relevant entries from \tm, and select the best tool to solve individual tasks more accurately. 
We evaluate \tm on learning varied text generation and text-to-image generation neural tools. Compared to no-memory, generic agents, we find \tm-augmented agents predict tool performance 14.8\% and 28.7\% more accurately across text and multimodal generation scenarios. 
Moreover, \tm facilitates optimal tool selection among multiple choices by 21\% and 24\% absolute increases in respective scenarios.
\end{abstract}

\section{Introduction}

Recent advances in agents have drastically reshaped the landscape of generative tasks, especially by utilizing powerful tools supported by large language models (LLMs) \citep{schick2023toolformerlanguagemodelsteach} or vision-language models (VLMs) \citep{gao2025multimodalagenttuningbuilding,Carrasco_2025,clip}.
These agents supported with task-specific neural tools have demonstrated impressive capabilities across various downstream applications such as instruction following and reasoning in text modality \citep{kim2025biggenbenchprincipledbenchmark,ouyang2022training}, as well as visual understanding and editing in cross-modal scenarios \citep{shen2023hugginggptsolvingaitasks,wang2024genartistmultimodalllmagent, tang2025captionvideofinegrainedobjectcentric}.
Unlike traditional tools such as \texttt{calculator} or \texttt{sql\_executor} that always produce deterministic outputs for a given input, tools supported by neural models vastly expand the spectrum of tasks that cannot be tackled by deterministic tools, such as answering ad-hoc questions \citep{chen2017readingwikipediaansweropendomain,karpukhin2020densepassageretrievalopendomain} or search relevant news on the web \citep{nakano2022webgptbrowserassistedquestionansweringhuman,dunn2017searchqanewqadataset}. Nonetheless, what are brought along with this wider applicability of neural tools is the uncertainty of their outputs---there is no guarantee that a \texttt{QA} tool can answer an arbitrary question correctly, especially when the answer to that question is open-ended \citep{lewis2021retrievalaugmentedgenerationknowledgeintensivenlp} or changes over time \citep{guu2020realmretrievalaugmentedlanguagemodel}.

\begin{figure}[t]
\centering
\includegraphics[width=\linewidth]{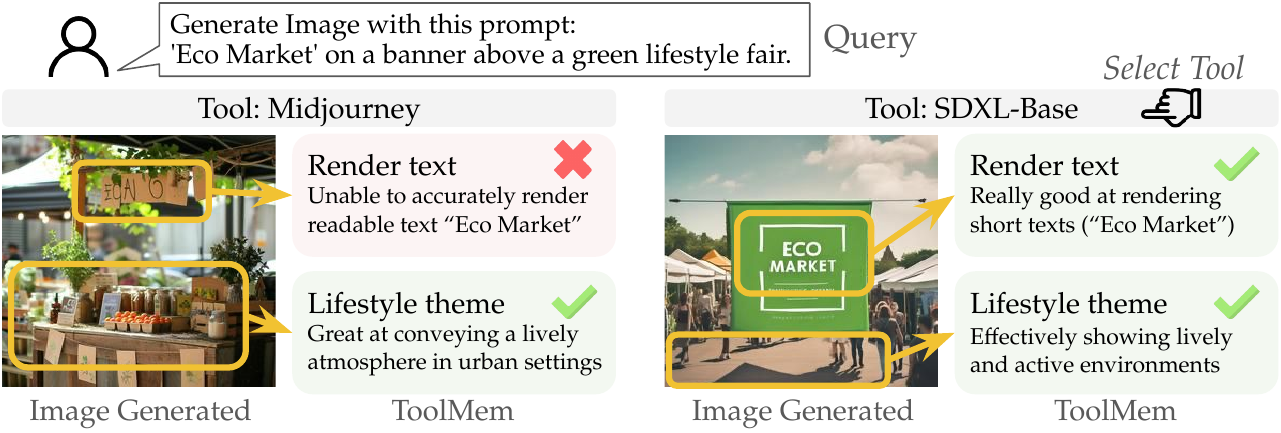}
\caption{Example of identifying better-performing tools based on learned \tm. For example, compared to the image generation tool instantiated by {\it Midjourney}, {\it SDXL-Base} tool is better at rendering short text such as the queried ``Eco Market''. We thus store this information in \tm and prioritize selecting {\it SDXL-base} in later tasks that ask to render short texts.}
\label{fig:teaser}
\vspace{-2mm}
\end{figure}

Moreover, current agents typically solve tasks using a pre-designated set of tools by human experts, and only get to learn to use these tools at test time by contextualizing on static textual descriptions for tool functionalities \citep{qin2023toolllm,guo2024stabletoolbench}. More often than not, the agent may be provided with multiple tools possessing similar functionalities \citep{tang2023toolalpaca,li2023api}, thus making it harder to distinguish the best tool to use for a certain task, because the agent has no prior knowledge about the individual expertise of these seemingly similar tools \citep{yao2023reactsynergizingreasoningacting}. Taking \autoref{fig:teaser} as an example, the two text-to-image tools have different expertise: while both might be good at representing lively environment, the later one may excel at rendering text. Given the task of generating image with text, the agent better pick the later tool to ensure better performance. 

Human gradually learn to distinguish and select between tools by interacting with tools and gathering knowledge about their individual properties through past experiences. 
Recent methods have explored some techniques to gather explicit knowledge from experiences by reflecting on \citep{shinn2023reflexion} and refining \citep{madaan2023selfrefineiterativerefinementselffeedback} generated solutions, or constructing short-term workflows \citep{wang2024agentworkflowmemory} and long-term memory \citep{xu2025amemagenticmemoryllm}, and even curating new tools to assist target downstream tasks \citep{wang2023voyageropenendedembodiedagent,wang2024troveinducingverifiableefficient}. 
Besides explicitly storing the knowledge, approaches like Toolformer \citep{schick2023toolformerlanguagemodelsteach} and ToolLLM \citep{qin2023toolllmfacilitatinglargelanguage} also explored parametric updates via self‑supervised fine‑tuning and instruction tuning on tool-using experiences.
However, these methods focus on learning task-specific procedures with a fixed tool, instead of distinguishing between a group of functionally similar tools to optimize downstream tool selection and performance. More concretely, agents lack a mechanism to build and update an internal, dynamic memory that encapsulates the strengths and weaknesses of diverse generative tools.

To bridge this gap, we introduce \tm, a framework that empowers agents to learn and apply tool-specific capability memories (\S\ref{sec:2:toolmem}). \tm is built upon three core components: (1) a structured capability memory initialized with a taxonomy that categorizes tool behaviors by proficiency levels, enabling consistent updates and targeted retrieval; (2) a feedback generation process that evaluates tool outputs using either human annotations or automated metrics—such as LLM-based judgment scores—to extract fine-grained insights into tool performance; and (3) a dynamic memory update mechanism that incorporates new experiences through retrieval-augmented generation (RAG), allowing the agent to refine, revise, and expand its tool knowledge over time. During inference, \tm-augmented agents retrieve relevant memory entries based on the current task and inject them into the input context, enabling more accurate tool selection and solution generation.

We first study: \textit{Can agents learn and estimate tool capabilities through interaction?}  (\S\ref{sec:3:perf-predict})
We construct \tm about image generation tools on \gb \citep{li2024genaibenchevaluatingimprovingcompositional} and various text-oriented tools on \bb \citep{kim2025biggenbenchprincipledbenchmark}, then evaluate if \tm-augmented agent can accurately estimate the performance of these tools. We find that \tm can predict solution quality using varied tools 14.8--28.7\% more accurately than \gr agent without prior knowledge about tools.

Next, we ask: \textit{Can agents leverage \tm to optimize performance through tool selection?} (\S\ref{sec:4:advanced-scenarios})
We examine \tm-augmented agents on selecting the best-performing tool for individual tasks among multiple tool candidates, and find it outperforming \gr agent approach by 21\% and 24\% on \gb and \bb benchmarks (\S\ref{sec:4:advanced-scenarios}).

In short, our work builds memory-adaptive tool-using agents, and uniquely addresses the dynamic nature of neural tools by enabling agents to grasp an evolving understanding of accessible tools.
\section{\tm: Learning Tool Capability Memory}
\label{sec:2:toolmem}

In this section, we introduce our \tm framework that initializes a structured memory (\S\ref{sec:2.1:mem-init}), learns and updates tool capability knowledge through experiences (\S\ref{sec:2.2:learn-toolmem}), and eventually solves new tasks according to retrieved memory entries (\S\ref{sec:2.3:task-solving}).
Overall, \tm equips agents to build tool capability memories progressively learned from past tool usage experiences, and leverages this dynamic knowledge repository to enhance its decision-making.

\subsection{Structured Memory Initialization}
\label{sec:2.1:mem-init}
The evolution of \tm is an iterative process that builds upon previous states. As such, the initial memory plays a critical role in guiding subsequent developments. 

Our goal is for agents $\pi$ to learn the capabilities of a set of tools $T = \{t\}$ with similar functionalities to solve a particular task (e.g., text-to-image generation, instruction following), where each tool $t$ is instantiated with a neural model (e.g., Midjourney, GPT). 
For each tool $t$, we initialize its capability memory $\mathcal{M}_t = \emptyset$ and aim to collect a set of knowledge entries $M_t = \{m_t\}$, each describing some properties of tool behaviors in natural language (NL).
While a tool $t$ can be good at certain scenarios and bad at others, we categorize memory entries based on the varied proficiency levels the agent possesses, namely $C$ = $\{${\it proficient at} ($^p$), {\it good at} ($^g$), {\it bad at} ($^b$), {\it weak at} ($^w$) $\}$. 
Moreover, these proficiency-aware categories can be associated with concrete numerical measures of tool performance, $+2$, $+1$, $-1$, $-2$, when more accurate measures are useful. We thus denote \tm as the union of all memory categories $\mathcal{M} = \cup_{c \in C} \mathcal{M}^{c}$. This structured memory initialization offers flexibility in memory update and retrieval, which we will provide more details on next.

\subsection{Learning Tool Capabilities from Experiences}
\label{sec:2.2:learn-toolmem}

\begin{figure}[t]
\centering
\includegraphics[width=\linewidth]{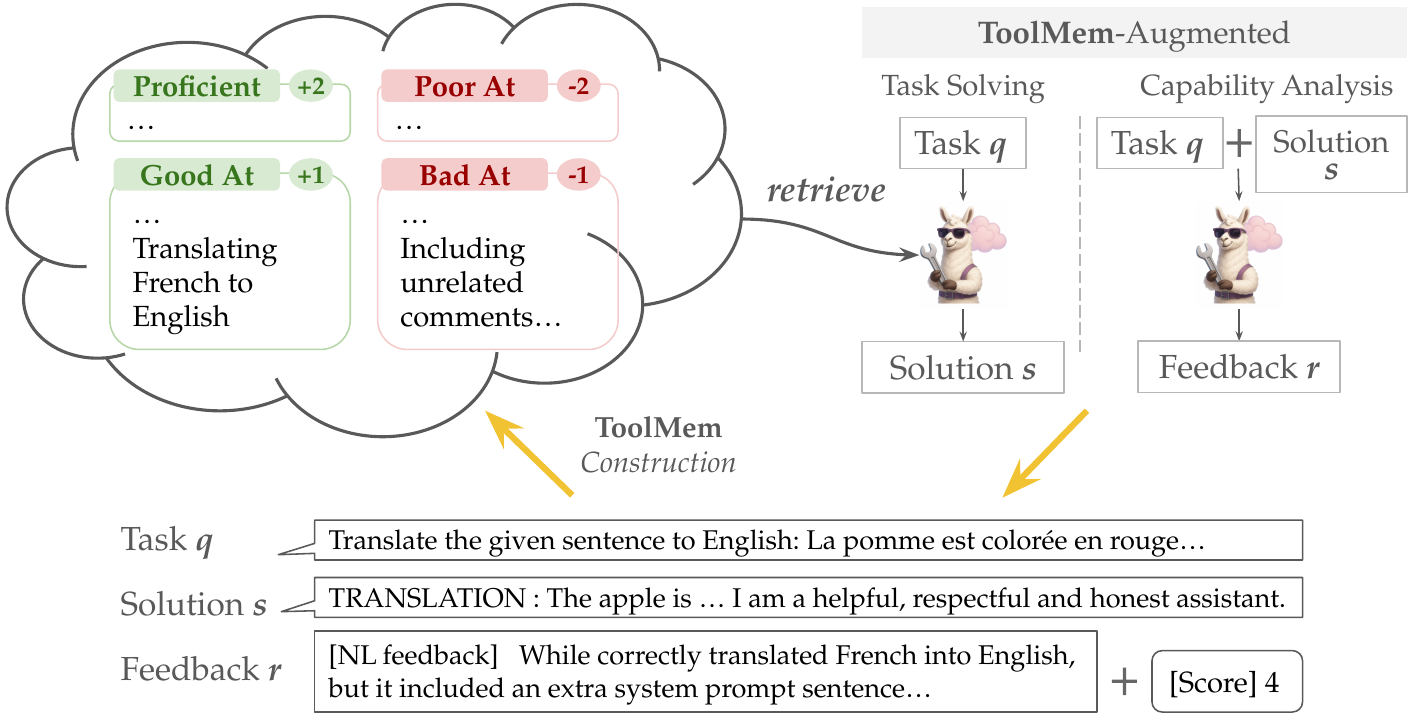}
\caption{Illustration of the \tm construction process from past experiences (left), as well as \tm-augmented task-solving and capability analysis at downstream (right).}
\label{fig:feedback_generation}
\end{figure}

\noindent \textbf{Constructing Memory from Experiences} \quad
Agents learn more about the tool capabilities by interacting with them in task-solving experiences. 
More concretely, the agent begins by receiving a task $q$ (e.g., an NL instruction), as well as a corresponding solution $s_t$ using a designated tool $t$ (e.g., an image generated by a text-to-image tool $t$, an answer generated by a text generation tool $t$). 
As an informative learning signal of this task, a reward system $R$ provides feedback $r_t = R(q, s_t)$ on the quality of the solution $s_t$ in tackling the task $q$, which could be numerical quality measures (e.g., 1-5 Likert scale) or NL feedback (as exemplified in \autoref{fig:feedback_generation}).
Together, the task $q$, tool-using solution $s_t$, and its quality feedback $r_t$ form an experience $e_t = (q, s_t, r_t)$ from interaction with tool $t$, upon which we can construct capability memories about.
In our study, we focus on building \tm from experiences annotated by human experts for image generation tools and by llm-as-a-judge for text generation tools. These annotations are provided as part of the training datasets, typically accompanied by high-quality solutions and reliable feedback.
We introduce a memory induction module $I_{\it LM}$ instantiated with an LM, to summarize the capabilities of tool $t$ from a pertaining experience $e_t$ via $I_{\it LM}(e_t) \rightarrow m_t$.

\noindent \textbf{Updating Agent Memory} \quad
Given a new memory entry $m$ derived from an experience $e$, a naive memory update would directly augment the existing memory repository by $\mathcal{M} \leftarrow \mathcal{M} \cup \{m\}$. However, this simplistic approach often introduces redundancy and inflates computational overhead. Specifically, entries that are highly similar or overlapping may accumulate without being appropriately consolidated, and prior incomplete or suboptimal entries might persist without receiving necessary updates.

To enable flexible memory refinement capabilities, we introduce a structured and refined memory updating procedure. Formally, we denote the agent's current memory as
$\mathcal{M} = \cup_{c \in C}\mathcal{M}^c$
partitioned into $C$ categories introduced in \S\ref{sec:2.1:mem-init}. Given the newly constructed experience $e = (q, s, r)$, we first perform a targeted retrieval within each memory category $\mathcal{M}^{c}$ to identify the top-$k$ most semantically relevant memory entries.
$\mathcal{M}^{c}_{\text{retrieved}} = \textit{Retriever}(e, \mathcal{M}^{c}), \forall c \in C$ where $\textit{Retriever}(\cdot, \cdot)$ calculates some similarity measure (e.g., cosine similarity) computed between embedded memory entries.

Next, we construct a condensed contextual set from these retrieved entries across all categories:
$\mathcal{M}_{\text{retrieved}} = \bigcup_{c \in C} \mathcal{M}^{c}_{\text{retrieved}}$.
Subsequently, the condensed contextual set $\mathcal{M}_{\text{context}}$ along with the new experience $e = (q, s, r)$ is passed into the \tm induction module $I$ to for entry refinement
$\mathcal{M}_{\text{updated}} = I(\mathcal{M}_{\text{retrieved}}, e)$ then memory update $\mathcal{M} \leftarrow (\mathcal{M} \setminus \mathcal{M}_{\text{context}}) \cup \mathcal{M}_{\text{updated}}$
The refinement process involves adding novel insights, updating incomplete entries, removing redundant information, and merging semantically related entries (see exact prompts in \S\ref{app:prompts}.

\subsection{Solving Tasks with Retrieved Memory}
\label{sec:2.3:task-solving}

Analogous to how humans draw upon past experiences when confronted with a new task, for a given task $q'$ at test time, our task-solving agents can (i) retrieve top-$k$ relevant memory entries from \tm utilizing a retriever model ${\it Retriever}(q', \mathcal{M}) \rightarrow \mathcal{M}_{q'}$, and then (ii) augmented these relevant memory entries to input contexts to generate solutions $\pi(q', \mathcal{M}_{q'}) \rightarrow s'$ or estimate tool performance $\pi(q', \mathcal{M}_{q'}) \rightarrow r'$. As \tm contains multiple categories $\mathcal{M} = \cup_{c} \mathcal{M}^c$, we separately retrieve top-$k$ relevant entries from each category $\mathcal{M}^c$ and provide all of them in context.
This RAG strategy addresses two key challenges: (1) relevance filtering: focusing on contextually useful entries and discarding irrelevant ones; (2) scalability: preventing computation overload by restricting retrieval to a small number of entries per category.

\section{Experiments: Tool Performance Prediction}
\label{sec:3:perf-predict}

In this section, we first introduce the baseline approaches in comparison to our \tm (\S\ref{sec:3.1:baselines}), then demonstrate \tm's effectiveness on two representative tasks --- text generation (\S\ref{sec:3.4:text-gen}) and text-to-image generation evaluation (\S\ref{sec:3.3:image-eval}).
We focus on \tm usage under the {\it performance prediction} scenario in this section. Later in \S\ref{sec:4:advanced-scenarios}, we further demonstrate its utility when extrapolating to \textit{model selection} under accuracy and efficiency considerations.

\subsection{Baseline and \tm Approaches}
\label{sec:3.1:baselines}

We introduce the two baseline approaches and our \tm method for experimentation.

\noindent \textbf{Generic Agent} \quad
Our first baseline is a vanilla agent without tool-specific memory beyond general information (e.g., tool's name, basic skills, input/output) at inference. This setup is similar to the current tool-selection approach adopted by state-of-the-art generative agents \citep{wang2024genartistmultimodalllmagent}. Agents must rely heavily on the parametric knowledge acquired through the training of their backbone LM to estimate tool capabilities. 

\noindent \textbf{Few-Shot Memory} \quad
Without the need to induce \tm, we propose another stronger baseline on top of \gr that also retrieves top-12 training examples, i.e., (task, score, rubric) triplets, that are most relevant to the test example at hand. 
We denote this the \fs setting and compare it to our \tm-augmented LM, to show the effectiveness of derived capability memory against implicit information from raw examples.

\noindent \textbf{\tm} \quad
Instead of employing memory with generic or raw experiences, our \tm contains carefully induced and updated memory from past experiences. Particularly, we induce memory entries from the same set of training examples used for retrieval in \fs setting, and test \tm-augmented LM on the same test set.

For generations, we use GPT-4o (the \texttt{gpt-4o-mini-2024-07-18} checkpoint) for all approaches by default; we use a temperature $t = 0.0$ and sample $n=1$ output.
For retrieval, we use OpenAI’s \texttt{text-embedding-ada-002}\footnote{\url{https://platform.openai.com/docs/models/text-embedding-ada-002}} model to generate embeddings and implement RAG using ChromaDB\footnote{\url{https://www.trychroma.com/}} that measures task-memory similarity using (negative) cosine distance.
During \tm construction, we retrieve $k=6$ entries from each memory category and perform memory update on them.
At test time, the agent retrieves $k=12$ entries from each memory category and grounds on them to generate task solutions; this setting scores the best in our ablation study on $k$ value choices in \S\ref{app:topK}.

\subsection{Text Generation}
\label{sec:3.4:text-gen}

\paragraph{Dataset: \bb}
We evaluate \tm on diverse text-generation tasks from the \bb benchmark \citep{kim2025biggenbenchprincipledbenchmark}, which includes 696 task instances spanning 8 capability dimensions, such as reasoning, safety, and tool usage, evaluated across 103 language models. We split the data into 211 training and 485 testing examples for our experiments.

Each task $q$ consists of a natural language instruction composed of a system prompt and user input. The solution $s_t$ is the textual response generated by a specific language model (tool) $t$. The quality feedback $r_t = R(q, s_t)$ includes a numerical score (on a 5-point Likert scale), a detailed score rubric, and NL feedback by GPT-4. We use feedback generated by GPT-4 to construct \tm for each tool, which are then used to improve downstream score prediction accuracy for new prompts.

We focus on six language model tools with varying capabilities and accessibility levels:
(1) Proprietary models: \textit{Claude-3-Sonnet}, \textit{GPT-3.5-Turbo};\footnote{Specific checkpoints are: {\tt claude-3-sonnet-20240229}, {\tt gpt-3.5-turbo-1106}.}
(2) Top-tier open-source models: \textit{Meta-Llama-3-70B-Instruct}, \textit{Qwen-110B-Chat};
(3) Smaller models: the strongest sub-7B model, \textit{Gemma-1.1-2B-It}, and the smallest chat-capable model, \textit{Qwen1.5-0.5B-Chat}.

\paragraph{Evaluation: Score Prediction}
We evaluate \tm-augmented LMs under a score prediction evaluation setup on how accurately they measure the tool generation quality. 
In our vanilla evaluation setup, each method is tasked with predicting a quality score for the generated image or text given a task prompt.
The goal is to predict the quality score of a prompt-response pair $q$, using the task-specific rubrics provided in the dataset, also rated on a 1--5 Likert scale. We take the \texttt{gpt4-04-turbo-score} as ground-truth, as no human scores are provided.

We measure the consistency of predicted scores $s_{\it pred}$ to the ground-truth scores $s$ by 
(i) the average mean absolute error (MAE) across all $N$ test examples ${\it MAE} = \frac{1}{N} \sum_{i=1}^{N} |s^{(i)}_{\it pred} - s^{(i)}|$, 
(ii) the root mean squared error by ${\it RMSE} = \sqrt{\frac{1}{N} \sum_{i=1}^{N} (s^{(i)}_{\it pred} - s^{(i)})^2}$, 
(iii) the Pearson correlation coefficient between predicted and true scores:
${\it Pearson} = \frac{\sum_{i=1}^{N} (s^{(i)}_{\it pred} - \bar{s}_{\it pred})(s^{(i)} - \bar{s})}{\sqrt{\sum_{i=1}^{N} (s^{(i)}_{\it pred} - \bar{s}_{\it pred})^2} \sqrt{\sum_{i=1}^{N} (s^{(i)} - \bar{s})^2}}$, where $\bar{s}_{\it pred}$ and $\bar{s}$ are the means of the predicted and ground-truth scores, respectively.


\paragraph{Results and Analysis}
As shown in \autoref{table:qa_score_full_comparison}, \tm reduces MAE by 14.8\% and RMSE by 14.5\% on average than \gr baseline;
and improves ranking fidelity—Pearson correlation increases by 76.7\%.

\noindent \textit{Tools with varied sizes} \quad  
For the two smallest and least capable tools (\textit{Qwen1.5-0.5B-Chat} and \textit{gemma-1.1-2B-it}), \tm boosts Pearson correlation significantly, lifting it from near-random (${-}\!0.007$ and $0.120$) to $0.405$ and $0.324$, highlighting the valuable, tool-specific priors that are initially lack for these lightweight models and are augmented through \tm.  
Even for the strongest open-weight tool (\textit{Meta-Llama-3-70B-Instruct}), \tm achieves a notable 38.9\% increase in Pearson correlation, from $0.175$ to $0.243$, while slightly reducing RMSE.

\begin{wraptable}[24]{r}{0.68\textwidth}
\vspace{-6mm}
\centering
\small
\caption{
Performance prediction of text-generation tools.
}
\vspace{2mm}
\label{table:qa_score_full_comparison}
\resizebox{0.68\textwidth}{!}{%
\begin{tabular}{ll|ccc}
\toprule
\multicolumn{1}{c}{\bf Tool} & \multicolumn{1}{c|}{\bf Method} & \bf MAE $\downarrow$ & \bf RMSE $\downarrow$ & \bf Pearson $\uparrow$ \\
\midrule
\multirow{3}{*}{Llama3 70B}
 & \gr & \textbf{0.571} & 1.036 & 0.175 \\
 & \fs & 1.115 & 1.566 & 0.163 \\
 & \tm  & 0.610 & \textbf{1.025} & \textbf{0.243} \\
\midrule
\multirow{3}{*}{Qwen 110B}
 & \gr & 0.602 & 0.991 & 0.228 \\
 & \fs & 1.155 & 1.605 & 0.146 \\
 & \tm  & \textbf{0.600} & \textbf{0.988} & \textbf{0.246} \\
\midrule
\multirow{3}{*}{Gemma 2B}
 & \gr & 1.004 & 1.528 & 0.120 \\
 & \fs & 1.008 & 1.429 & 0.271 \\
 & \tm  & \textbf{0.932} & \textbf{1.252} & \textbf{0.324} \\
\midrule
\multirow{3}{*}{Qwen1.5 0.5B}
 & \gr & 1.769 & 2.136 & -0.007 \\
 & \fs & 1.381 & 1.762 & 0.102 \\
 & \tm  & \textbf{1.080} & \textbf{1.400} & \textbf{0.405} \\
\midrule
\multirow{3}{*}{Claude3}
 & \gr & \textbf{0.546} & 0.983 & \textbf{0.313} \\
 & \fs & 1.085 & 1.555 & 0.140 \\
 & \tm  & 0.551 & \textbf{0.979} & 0.310 \\
\midrule
\multirow{3}{*}{GPT-3.5}
 & \gr & 0.625 & 1.095 & 0.224 \\
 & \fs & 0.924 & 1.392 & 0.204 \\
 & \tm  & \textbf{0.588} & \textbf{0.998} & \textbf{0.332} \\
\bottomrule
\end{tabular}
}
\vspace{0mm}
\end{wraptable}

\noindent \textit{Top closed tools have diminishing returns} \quad  
For \textit{gpt-3.5-turbo-1106} and \textit{claude-3-sonnet}, baseline predictions already achieve relatively strong Pearson correlations ($\geq 0.224$), while \tm still manages to enhance correlation for GPT by 48.2\%.



\noindent \textit{Instability of Raw Experiences} \quad
Across all tools, the \fs strategy underperforms \gr across metrics and often doubles the error for the two top models (e.g., MAE $+95\%$ on Meta-Llama-3-70B).  
In contrast, \tm delivers robust improvements without such regressions, underscoring the value of a compact, induced capability memory over raw example reference.

\subsection{Text-to-Image Evaluation}
\label{sec:3.3:image-eval}

\paragraph{Dataset: \gb}
We evaluate text-to-image generation tools using the \gb benchmark \citep{li2024genaibenchevaluatingimprovingcompositional}, which contains 1,600 diverse natural language instructions spanning compositional scenarios such as object counting, spatial relations, and negation. 
We randomly sample 200 training and 800 testing examples for experiments.

Each instruction is paired with six images generated by different tools, including \textit{MidJourney}, \textit{DALL{\small \textperiodcentered}E~3}, \textit{SDXL Turbo}, \textit{DeepFloyd I-XL v1}, \textit{SDXL 2.1}, and \textit{SDXL Base}.
Each text-image pair is annotated by three human raters on a 1–5 Likert scale, reflecting how well the image aligns with the prompt. As multiple human annotators can introduce variance in prediction scores, we take the first annotator's rating as the ground-truth score to ensure label consistency. 
Our goal is to predict the quality score $r_t$ based on the task input $q$ (image generation prompt) and the solution $s_t$ (the generated image using tool $t$) with agents having no memory, few-shot examples, or \tm.

\paragraph{Evaluation: Score \& Description Prediction}
We similarly evaluate \tm-augmented LM to predict the alignment score of a prompt-image pair $q$, rated on a 5-point Likert scale. 

While score prediction may be subject to individual human judgment and less robust, we alternatively evaluate image quality through text-image alignment: given a text prompt $q$ (e.g. ``Three birds flying in the sky''), we ask agent to predict a text description $d_{\it pred}$ for the image $s^{\it image}$ generated by tool $t$. We then evaluate the alignment between the image and the predicted description using the VQA score \citep{li2024genaibenchevaluatingimprovingcompositional}, computed as $s_{\it vqa} = {\it VQA}(s^{\it image}, d_{\it pred})$. A higher VQA score indicates stronger alignment between the image and the textual description. We report average VQA Score in \autoref{table:score-pred-basic}.

\paragraph{Results and Analysis}
\autoref{table:score-pred-basic} lists the six image–generation tools in descending order of capability.
Overall, \tm consistently improves over the \gr baseline by decreasing MAE by 28.7\% and RMSE by 26.6\% across tools.
The gains, however, are not uniform:

\noindent \textit{Top-tier closed models (DALLE 3, Midjourney 6).} \quad
\tm beats the \gr agent (–7.4\% and –12.2\% MAE, respectively) but trails \fs by 3.7\% to 6.7\%. This suggests that when the underlying generator is highly capable, large in-context exemplars are sufficient and an external capability memory brings limited extra benefit. 
However, \tm delivers consistent gains in VQA scores, with improvements of +2.1\% for DALLE 3 and +2.0\% for Midjourney 6. These steady gains highlight \tm's ability to fine-tune descriptions even for top-performing models, effectively capturing subtle yet crucial details that baseline methods might overlook.

\noindent \textit{Mid/low-tier open models (DeepFloyd\_I\_XL\_v1, SDXL-Base, SDXL-Turbo, SD-2-1).} \quad
Here \tm clearly shows its strength: it achieves the lowest MAE and RMSE on all four models, reducing MAE by 26.1–42.6\% and RMSE by 25.5–36.6\% relative to \gr, and outperforming \fs by 6.7–18.0\% MAE.

\begin{wraptable}[24]{r}{0.7\textwidth}
\vspace{-7mm}
\centering
\small
\caption{Predicting text-to-image tool performance on \gb.}
\label{table:score-pred-basic}
\vspace{1mm}
\begin{tabular}{ll|ccc}
\toprule
\multicolumn{1}{c}{\multirow{2}{*}{\bf Tool}} & \multicolumn{1}{c|}{\multirow{2}{*}{\bf Method}} & \multicolumn{2}{c}{\bf Score Prediction} & \multicolumn{1}{c}{\bf Desc. Pred.} \\
\cmidrule(lr){3-4} \cmidrule(lr){5-5}
 & & \bf MAE $\downarrow$ & \bf RMSE $\downarrow$ & \bf VQA Score (\%) $\uparrow$ \\
\midrule
\multirow{3}{*}{DALLE 3} 
 & \gr & 1.009 & 1.405 & 81.30 \\
 & \fs & \textbf{0.875} & \textbf{1.180} & 81.64 \\
 & \tm & 0.934 & 1.204 & \textbf{82.98} \\
\midrule
\multirow{3}{*}{Midjourney 6} 
 & \gr & 1.085 & 1.434 & 81.19 \\
 & \fs & \textbf{0.919} & 1.227 & 82.21 \\
 & \tm & 0.953 & \textbf{1.212} & \textbf{82.84} \\
\midrule
\multirow{3}{*}{DeepFloyd} 
 & \gr & 1.334 & 1.703 & 79.17 \\
 & \fs & 1.058 & 1.393 & 79.74 \\
 & \tm & \textbf{0.986} & \textbf{1.268} & \textbf{81.30} \\
\midrule
\multirow{3}{*}{SDXL-Base} 
 & \gr & 1.481 & 1.841 & 80.47 \\
 & \fs & 1.111 & 1.422 & 81.37 \\
 & \tm & \textbf{0.978} & \textbf{1.285} & \textbf{82.03} \\
\midrule
\multirow{3}{*}{SDXL-Turbo} 
 & \gr & 1.566 & 1.932 & 79.22 \\
 & \fs & 1.101 & 1.400 & 80.01 \\
 & \tm & \textbf{0.996} & \textbf{1.331} & \textbf{81.73} \\
\midrule
\multirow{3}{*}{SDXL-2-1} 
 & \gr & 1.645 & 1.981 & 75.70 \\
 & \fs & 1.151 & 1.477 & 75.91 \\
 & \tm & \textbf{0.944} & \textbf{1.255} & \textbf{78.90} \\
\bottomrule
\end{tabular}
\vspace{-2mm}
\end{wraptable}

\tm also exhibits significant improvements for VQA scores, particularly in mid- and low-tier models where gains reach +3.17\% for SDXL-Turbo and +4.23\% for SDXL-2-1. This pattern reflects \tm's ability to adapt to each model's specific strengths and weaknesses, resulting in more accurate and context-aware descriptions. The learnable memory is especially effective for weaker tools that lack robust prior knowledge, helping narrow the gap between less capable generators and their stronger counterparts.

In summary, \tm offers the greatest benefit where it is most needed—narrowing the gap between weaker generators and their stronger peers—while remaining competitive on state-of-the-art closed models. Its dual effectiveness in both score prediction and precise description prediction highlights its capacity to enhance model evaluation by leveraging learned memory, ultimately providing higher VQA scores and more precise model-specific descriptions.

\section{Experiments: Performant Tool Selection}
\label{sec:4:advanced-scenarios}



In addition to predicting a single tool's performance, we can select the better-performing tool among multiple choices based on the \tm learned for these tools. 
We construct \tm and evaluate on the same datasets on {\sc GenAI-Bench} and {\sc BigGen-Bench}, and compare the same three methods (\gr, \fs, \tm) as in \S\ref{sec:3:perf-predict}.

\paragraph{Evaluation Metrics}
Let $s_A$ and $s_B$ denote the ground-truth scores of tools A and B, and $s_{\it pred,A}$ and $s_{\it pred,B}$ be their predicted scores. Since for equal pairs selecting either tool is equally acceptable, we evaluate tool selection performance over the subset of \emph{unequal} pairs $D = \{i: s_A^i \ne s_B^i\}$. For each $i \in D$, we define $TP_< = |\{i \in D: s_A^i < s_B^i \wedge s_{\it pred,A}^i < s_{\it pred,B}^i\}|$, $P_< = |\{i \in D: s_{\it pred,A}^i < s_{\it pred,B}^i\}|$, and $R_< = |\{i \in D: s_A^i < s_B^i\}|$. We define $TP_>$, $P_>$, and $R_>$ analogously for the “$>$” case. Then we compute $F_{1<} = \frac{2 TP_<}{P_< + R_<}$, $F_{1>} = \frac{2 TP_>}{P_> + R_>}$, and $\mathrm{Acc} = \frac{TP_< + TP_>}{|D|}$, where $F_{1<}$ (resp.\ $F_{1>}$) is the F$_1$ score for predicting $s_A < s_B$ (resp.\ $s_A > s_B$) over the non-equal set $D$, and Acc is the proportion of cases where the predicted and ground-truth preference directions match.


\paragraph{Text Generation Evaluation}

On \bb, we evaluate six tool pairs as listed in \autoref{table:model-selection-textgen}.
The selected pairs include comparisons between top open-source and closed-source models (e.g., Meta-Llama-3-70B vs.\ Claude-3-Sonnet), similarly strong closed models (Claude-3-Sonnet vs.\ GPT-3.5-Turbo), and strong open models (Meta-Llama-3-70B vs.\ Qwen110B), models from the same family with large capacity gaps (Qwen110B vs.\ Qwen0.5B), and small-scale model comparisons (e.g., Gemma-2B vs.\ Qwen0.5B).
These combinations cover diverse settings across model families, sizes, and access types.

As shown in \autoref{table:model-selection-textgen}, we find that \tm offers clear and consistent accuracy advantages over \gr and \fs settings by 21\% and 18\% on average. 
\tm increases the F$_{1<}$ score, which measures the ability to identify the cases where weaker tool excels, increases from 0.00 to 0.28 from \gr in the Meta-Llama-3-70B vs.\ Claude-3-Sonnet comparison, on average it improves from \gr's 0.03 to 0.09.
F$_{1>}$ improves more significantly from 0.12 to 0.39 on average, with \tm achieving the highest score on five of the six tool pairs. 
\tm boosts performance most substantially when tool pairs exhibit large capability gaps, e.g., \tm improves accuracy by +0.55 for Qwen110B vs.\ Qwen0.5B and +0.28 for GPT-3.5-Turbo vs.\ Gemma-2B. 
Even for tool pairs with smaller capability gaps (e.g., Meta-Llama-3-70B vs.\ Claude-3-Sonnet), \tm maintains an edge over both baselines.

\begin{table}[t!]
\centering
\small
\caption{Selecting better text generation tools. Best results for each tool pair are bold.}
\label{table:model-selection-textgen}
\begin{tabular}{ll|ccc|ccc|ccc}
\toprule
\multicolumn{2}{c|}{\multirow{2}{*}{\bf Tool Pair}} & \multicolumn{3}{c}{\bf \gr} & \multicolumn{3}{c}{\bf \fs} & \multicolumn{3}{c}{\bf \tm} \\
{} & {} & F$_{1<}$ & F$_{1>}$ & Acc. & F$_{1<}$ & F$_{1>}$ & Acc. & F$_{1<}$ & F$_{1>}$ & Acc. \\
\midrule
Llama3 & Claude3 & 0.00 & 0.09 & 0.03 & 0.20 & 0.06 & 0.07 & \textbf{0.28} & \textbf{0.10} & \textbf{0.11} \\
Claude3 & GPT-3.5-Turbo & 0.03 & 0.06 & 0.03 & \textbf{0.21} & 0.22 & 0.13 & 0.06 & \textbf{0.30} & \textbf{0.15} \\
Qwen 110B & Qwen 0.5B & 0.00 & 0.13 & 0.07 & 0.00 & 0.09 & 0.05 & 0.00 & \textbf{0.77} & \textbf{0.62} \\
Llama3 & Qwen 110B & 0.04 & 0.19 & 0.07 & 0.12 & \textbf{0.17} & 0.08 & \textbf{0.16} & 0.14 & \textbf{0.09} \\
Gemma 2B & Qwen 0.5B & 0.10 & 0.04 & 0.03 & 0.05 & 0.12 & 0.06 & \textbf{0.06} & \textbf{0.39} & \textbf{0.23} \\
GPT-3.5-Turbo & Gemma 2B & 0.00 & 0.24 & 0.11 & \textbf{0.08} & 0.27 & 0.14 & 0.00 & \textbf{0.62} & \textbf{0.39} \\
\midrule
\multicolumn{2}{c|}{\bf Average} & 0.03 & 0.12 & 0.06 & \textbf{0.11} & 0.16 & 0.09 & 0.09 & \textbf{0.39} & \textbf{0.27} \\
\bottomrule
\end{tabular}
\end{table}


\paragraph{Text-to-Image Generation Evaluation}
On \gb, we examine six representative tool pairs (in \autoref{table:model-selection-image-eval}) to test whether \tm-augmented LM can accurately identify the superior tool for individual tasks. More capable model is on the left side of each pair.
To showcase \tm utility in varied scenarios, we choose varied types of tool pairs. Specifically: 
(i) distinguishing comparably top-performing proprietary tools (DALL$\cdot$E 3, Midjourney) and against open-weight tools (DALL$\cdot$E 3, DeepFloyd), (Midjourney, DeepFloyd);
(ii) comparing tools with huge gaps from the same (SDXL-Base, SDXL-Turbo) and different model families (DALL$\cdot$E 3, SDXL-2.1).

\autoref{table:model-selection-image-eval} shows the consistent edge of \tm across varied tool pairs. Averaged over five tool pairs, accuracy on unequal pairs climbs from 0.09 (\gr) and 0.27 (\fs) to 0.33 with \tm, an absolute gain of +0.24 over \gr (+266\%) and +0.06 over \fs (+22\%). 

The hardest class F$_{1<}$, which measures the situation to find worse cases for a more capable model, improves most dramatically---rising from 0.09 to 0.32 compared to \gr and \fs approaches.
Meanwhile, F$_{1>}$ also edges up from 0.15 to 0.46 when augmented with \tm. 

The benefit brought by \tm shows more prominently as the capability gap of the tool pair grows. For instance, between comparable DALL$\cdot$E 3 and Midjourney tools, the gain of Acc. for \tm compare to \gr is relatively small (+0.12), yet comparing DALL$\cdot$E 3 to the weaker SDXL-2.1, \tm boosts accuracy from 0.08 to 0.53, achieves much larger gains.

Even within the same model family, \tm can effectively tell tools' capability differences, showcased by the +0.18 gain in the SDXL-Base vs.\ SDXL-Turbo comparison.
These results confirm that \tm consistently chooses better tools when the tool choices exhibit different properties.

\begin{table}[ht]
\small
\centering
\caption{Selecting better text-to-image generation tools. Best results for each tool pair are bold.}
\label{table:model-selection-image-eval}
\begin{tabular}{ll|ccc|ccc|ccc}
\toprule
\multicolumn{2}{c|}{\multirow{2}{*}{\bf Tool Pair}} & \multicolumn{3}{c|}{\bf \gr} & \multicolumn{3}{c|}{\bf \fs} & \multicolumn{3}{c}{\bf \tm} \\
{} & {} & F$_{1<}$ & F$_{1>}$ & Acc. & F$_{1<}$ & F$_{1>}$ & Acc. & F$_{1<}$ & F$_{1>}$ & Acc. \\
\midrule
DALLE 3 & DeepFloyd & 0.07 & 0.19 & 0.10 & 0.19 & \textbf{0.52} & 0.31 & \textbf{0.31} & \textbf{0.52} & \textbf{0.35} \\
DALLE 3 & Midjourney & 0.00 & 0.31 & 0.14 & 0.17 & 0.42 & 0.23 & \textbf{0.26} & \textbf{0.43} & \textbf{0.26} \\
DALLE 3 & SD-2-1 & 0.00 & 0.16 & 0.08 & 0.20 & 0.66 & 0.47 & \textbf{0.32} & \textbf{0.72} & \textbf{0.53} \\
SDXL-Base & SDXL-Turbo & 0.14 & 0.03 & 0.05 & 0.24 & \textbf{0.29} & 0.18 & \textbf{0.36} & 0.26 & \textbf{0.23} \\
Midjourney & DeepFloyd & 0.23 & 0.07 & 0.08 & 0.26 & 0.28 & 0.18 & \textbf{0.35} & \textbf{0.37} & \textbf{0.26} \\
\midrule
\multicolumn{2}{c|}{\bf Average} & 0.09 & 0.15 & 0.09 & 0.21 & 0.43 & 0.27 & \textbf{0.32} & \textbf{0.46} & \textbf{0.33} \\
\bottomrule
\end{tabular}
\vspace{-3mm}
\end{table}

\section{Related Work}

\paragraph{Tool‑Using Multimodal Agents}
Multimodal agents combine LLMs and LVLMs to solve complex tasks across modalities \citep{xie2024largemultimodalagentssurvey}. Early systems like Visual Programming \citep{gupta2022visualprogrammingcompositionalvisual} and ViperGPT \citep{surís2023vipergptvisualinferencepython} use LLMs to compose expert modules for visual reasoning. Visual ChatGPT \citep{wu2023visualchatgpttalkingdrawing} and HuggingGPT \citep{shen2023hugginggptsolvingaitasks} improve agent flexibility via prompt routing and tool selection from a VFM pool. Later methods such as Toolformer \citep{schick2023toolformerlanguagemodelsteach}, AssistGPT \citep{gao2023assistgptgeneralmultimodalassistant}, and ToolLLM \citep{qin2023toolllm} learn to invoke tools through self-supervision or in-context examples. While effective, these approaches assume fixed tool behaviors or rely on limited, static memory. 
In contrast, \tm enables adaptive, context-aware tool selection among choices, by building capability memories that capture tools' empirical strengths and weaknesses through experiences.

\paragraph{Generative AI Tools}
Unlike deterministic tools that produce predictable outputs, tools supported by generative neural models (e.g., \textit{DALLE-3}, \textit{GPT}) exhibit varied performances across task specifications \citep{tang2025generativeaicelanimationsurvey}. Some tools may excel at object counting, while others are better at spatial understanding \citep{sim2024evaluatinggenerationspatialrelations}. While prior work has relied on human expertise to navigate these nuances, current agent systems typically neglect such variance during tool selection \citep{wang2024genartistmultimodalllmagent}.
In contrast, \tm records empirical performance signals for each tool, enabling agents to make accuracy-aware decisions during task solving.

\paragraph{Memory-Augmented Agents}
Effective memory systems enable agents to operate over extended tasks. Foundational work like neural Turing machines \citep{graves2014neuralturingmachines} introduced external memory modules for querying past states. Recent systems such as Reflexion \citep{shinn2023reflexion} store in-session NL-formatted feedback as memory, while AWM \citep{wang2024agentworkflowmemory} captures reusable workflows across tasks. MemGPT \citep{packer2024memgptllmsoperatingsystems} and MemoryBank \citep{zhong2023memorybankenhancinglargelanguage} separate memory into short- and long-term slots with dynamic update strategies. While these methods enhance general reasoning and task procedures of agents, they do not model the behavior of external tools, especially those with functionalities unknown prior to testing. In contrast, \tm explicitly tracks tool-specific strengths and weaknesses, facilitating smarter tool selection for use.

\section{Conclusion and Future Work}

We introduced \tm, a closed-loop framework that equips multimodal agents with a learnable and evolving memory of tool capabilities. By integrating structured memory initialization based on a proficiency-level taxonomy, feedback-driven learning from LLM-generated critiques, and retrieval-augmented generation for memory refinement, \tm enables agents to continually improve their understanding of tool behaviors. Our experiments demonstrate that agents augmented with \tm achieve more accurate tool performance estimation and make better-informed tool choices, leading to improved outcomes in both text and image generation tasks.

This work opens up several promising directions for future exploration. One compelling avenue is extending ToolMem to support \textit{a wider array of generative and decision-making tools} across domains such as multimodal editing and code synthesis. 
Another direction is developing \textit{more advanced automated feedback} mechanisms to support autonomous memory construction. 
Additionally, analyzing the \textit{theoretical underpinnings of memory} maintenance can offer new insights into long-term memory scalability. 
Finally, incorporating \textit{human-in-the-loop} and \textit{principled memory consolidation} strategies may enhance learning from sparse or noisy experiences. 
We hope \tm can serve as a foundation for building adaptive agents that learn from prior interactions without repeated retraining, paving the way for more efficient, resilient, and generalizable AI systems.

\bibliographystyle{plainnat}
\bibliography{references}

\appendix

\newpage





\section{Analysis: Number of Memory Entries to Retrieve}
\label{app:topK}

We further conduct an in-depth analysis on top-$k$ retrieval granularity, by retrieving varying numbers of memory entries from \tm.

\begin{figure}[ht]
\centering
\includegraphics[width=\linewidth]{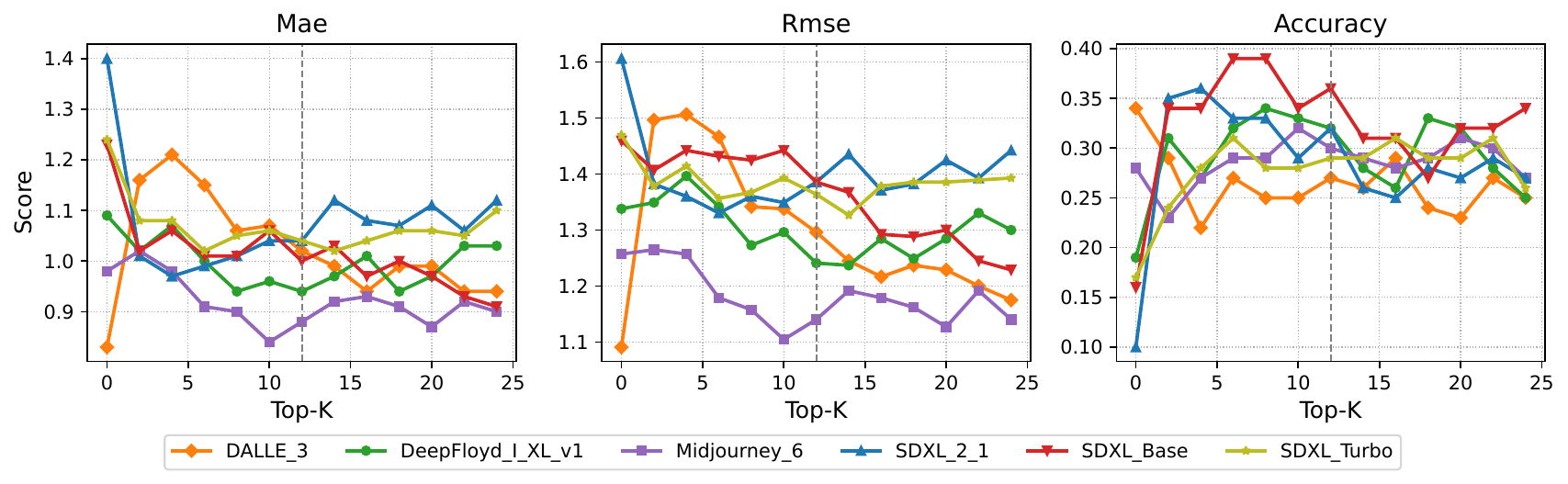}
\caption{\tm's score prediction accuracy on six text-to-image models across different top-$k$ memory retrieval sizes, evaluated on 100 samples. The dashed line at $k=12$ indicates most models' empirically optimal memory size.}
\label{fig:topK_trend}
\end{figure}

As shown in \autoref{fig:topK_trend}, 
We also compare performance when using varying \textit{top-k} values (0--24)\footnote{When \textit{top-k} = 0, predictions are made using only the tool name and its overview, without retrieving any memory items.}.
\autoref{fig:topK_trend} shows that larger \textit{top-k} values generally reduce MAE and RMSE but eventually plateau or degrade due to noise from excessive retrieval. Accuracy follows a similar trend, often peaking around \textit{top-k} = 10--14. For example, \tm’s predictions for Midjourney\_6 improve from an MAE of 0.98 at \textit{top-k} = 0 to 0.91 at \textit{top-k} = 18; SDXL\_Turbo’s MAE falls from 1.24 to 1.06 over the same range. These results suggest that leveraging a moderate number of high-relevance memory items helps balance predictive accuracy and computational efficiency.

\section{Prompt Templates}
\label{app:prompts}

\subsection{Text to Image Tools}


\begin{tcolorbox}[colback=gray!10, colframe=gray!60, coltitle=black,
title=Prompt for Feedback Generation, fonttitle=\bfseries]
The text prompt used to generate the image is ``\texttt{\{prompt\}}''. The Scoring metric used here is a 5-point Likert scale:

\textit{1 (Does not match at all)}: The generated image or video completely fails to align with the text prompt. It either lacks the required elements or depicts a scenario that contradicts the prompt.

\textit{2 (Has significant discrepancies)}: There are major mismatches between the text prompt and the visual output. Key elements may be missing, or relationships and attributes are significantly misrepresented.

\textit{3 (Has several minor discrepancies)}: The output aligns with the prompt in a general sense but has noticeable errors or omissions in key details. While the main elements are present, finer aspects are not accurately rendered.

\textit{4 (Has a few minor discrepancies)}: The output is mostly aligned with the prompt, with only a few small and less critical details being inaccurate or missing. The visual matches the description well but not perfectly.

\textit{5 (Matches exactly)}: The generated image or video perfectly matches the text prompt. All elements, relationships, attributes, and details align seamlessly with the description, with no noticeable discrepancies. 

This image scored \texttt{\{score\}} based on the human ratings. Now tell me why you think the image is or isn't following the text prompt.
\end{tcolorbox}


\newpage
\newpage
\begin{tcolorbox}[colback=gray!10, colframe=gray!60, coltitle=black,
title=Prompt for Memory Refinement, fonttitle=\bfseries, breakable]
You are an assistant responsible for maintaining an incremental memory of the tool's capabilities, limitations, and use cases. \\
You will receive a current memory and new feedback. Your job is to update the memory by integrating the new feedback without losing previously stored information. \\

\textbf{Rules:} \\
1. Start with the current memory. \\
2. Consider the new prompt and feedback, but do not simply repeat the feedback, think deep and get general insights. \\
3. Incrementally refine and update the memory to reflect any new insights. \\
4. Do not remove previously known capabilities unless the new feedback explicitly contradicts them. \\
5. If no new insights are provided, leave the memory unchanged. \\

\textbf{Instructions:} \\
Follow this procedure strictly: \\
- Review the current memory for model \texttt{\{tool\_name\}}: \\
\quad ``\texttt{\{current\_memory\}}'' \\
- Input task prompt to the tool was: \\
\quad ``\texttt{\{task\_prompt\}}'' \\
- Tool's answer was: \\
\quad ``\texttt{\{response\}}'' \\
- Score rubric was: \\
\quad ``\texttt{\{score\_rubric\}}'' \\
- The score for tool's response was: \\
\quad ``\texttt{\{score\}}'' \\
- Feedback for this answer received: \\
\quad ``\texttt{\{feedback\}}'' \\

Integrate the key insights of the tool's limitations and capabilities from the feedback into the existing memory. Use complete sentences starting with phrases like \texttt{good at}, \texttt{bad at}, \texttt{proficient at}, or \texttt{poor at} to describe your findings. \\
- If the feedback suggests new limitations, capabilities, or clarifications, add them. \\
- If the feedback suggests that something previously stated is incorrect, correct it. \\
- If the feedback suggests the tool struggles with a certain aspect (like counting multiple objects), note it. \\
- If no contradiction or new insight is provided, keep the memory as is. \\
- Think deeply and determine what the feedback actually means for the model's capability, rather than staying on the surface of a specific example. \\

In your final answer, output \textbf{ONLY} the updated overall memory of the tool. Do not include extra explanations. \\
\end{tcolorbox}

\begin{tcolorbox}[colback=gray!10, colframe=gray!60, coltitle=black,
title=Predict Image Description - \gr, fonttitle=\bfseries, breakable]
You are familiar with how the model \texttt{\{model\_name\}} generates images. \\
Your current memory of \texttt{\{model\_name\}} is: ``A text to image model'' \\

Based on this memory, predict how an image generated by \texttt{\{model\_name\}} for the following prompt would look: \\
Prompt: ``\texttt{\{prompt\}}'' \\

Describe the expected image in no more than 50 English words. Be concise and specific. \\
Only return the description text.
\end{tcolorbox}

\begin{tcolorbox}[colback=gray!10, colframe=gray!60, coltitle=black,
  title=Predict Image Description - \fs, fonttitle=\bfseries, breakable]
You are familiar with how the model \texttt{\{self.model\_name\}} generates images. \\
Your current memory of \texttt{\{self.model\_name\}} is: ``\texttt{\{current\_memory\}}'' \\

Here are some examples of the model's performance on specific prompts: \\
\texttt{\{few\_shot\_memory\}} \\

The scoring metric used here is a 5-point Likert scale: \\
\textit{1 (Does not match at all)}: The generated image or video completely fails to align with the text prompt. It either lacks the required elements or depicts a scenario that contradicts the prompt. \\
\textit{2 (Has significant discrepancies)}: There are major mismatches between the text prompt and the visual output. Key elements may be missing, or relationships and attributes are significantly misrepresented. \\
\textit{3 (Has several minor discrepancies)}: The output aligns with the prompt in a general sense but has noticeable errors or omissions in key details. While the main elements are present, finer aspects are not accurately rendered. \\
\textit{4 (Has a few minor discrepancies)}: The output is mostly aligned with the prompt, with only a few small and less critical details being inaccurate or missing. The visual matches the description well but not perfectly. \\
\textit{5 (Matches exactly)}: The generated image or video perfectly matches the text prompt. All elements, relationships, attributes, and details align seamlessly with the description, with no noticeable discrepancies. \\

Based on these examples and your memory of \texttt{\{self.model\_name\}}, predict how an image generated by \texttt{\{self.model\_name\}} for the following prompt would look: \\

Prompt: ``\texttt{\{prompt\}}'' \\

Describe the expected image in no more than 50 English words. Be concise and specific. \\
Only return the description text.
\end{tcolorbox}

\begin{tcolorbox}[colback=gray!10, colframe=gray!60, coltitle=black,
title=Predict Image Description - \tm, fonttitle=\bfseries, breakable]
You are familiar with how the model \texttt{\{model\_name\}} generates images. \\
Your current memory of \texttt{\{model\_name\}} is: ``\texttt{\{current\_memory\}}'' \\

Based on this memory, predict how an image generated by \texttt{\{model\_name\}} for the following prompt would look: \\
Prompt: ``\texttt{\{prompt\}}'' \\

Describe the expected image in no more than 50 English words. Be concise and specific. \\
Only return the description text.
\end{tcolorbox}









\begin{tcolorbox}[colback=gray!10, colframe=gray!60, coltitle=black,
title=Predict Score - \gr, fonttitle=\bfseries, breakable]
Your current memory (knowledge) of the tool \texttt{\{model\_name\}} is: ``A text to image model'' \\
Based on your memory of model \texttt{\{model\_name\}}, predict the score for the following prompt: ``\texttt{\{prompt\}}''. \\

The scoring metric used here is a 5-point Likert scale: \\
\textit{1 (Does not match at all)}: The generated image or video completely fails to align with the text prompt. It either lacks the required elements or depicts a scenario that contradicts the prompt. \\
\textit{2 (Has significant discrepancies)}: There are major mismatches between the text prompt and the visual output. Key elements may be missing, or relationships and attributes are significantly misrepresented. \\
\textit{3 (Has several minor discrepancies)}: The output aligns with the prompt in a general sense but has noticeable errors or omissions in key details. While the main elements are present, finer aspects are not accurately rendered. \\
\textit{4 (Has a few minor discrepancies)}: The output is mostly aligned with the prompt, with only a few small and less critical details being inaccurate or missing. The visual matches the description well but not perfectly. \\
\textit{5 (Matches exactly)}: The generated image or video perfectly matches the text prompt. All elements, relationships, attributes, and details align seamlessly with the description, with no noticeable discrepancies. \\

Now return a score between 1 and 5 for prompt \texttt{\{prompt\}}, based on your memory of the tool \texttt{\{model\_name\}}. Return a single number only, nothing else. \\
\end{tcolorbox}

\begin{tcolorbox}[colback=gray!10, colframe=gray!60, coltitle=black,
title= Predict Score - \fs, fonttitle=\bfseries]
Your current memory (knowledge) of the tool \texttt{\{self.model\_name\}} is: ``A text to image model''.

Here are some examples of the model's performance on specific prompts:
\texttt{\{samples\_prompt\}}.

Based on these examples and your memory of \texttt{\{self.model\_name\}}, predict the score this model will get for answering the following task prompt: ``\texttt{\{prompt\}}''.

The scoring metric used here is a 5-point Likert scale:

\textit{1 (Does not match at all)}: The model's response completely fails to align with the task prompt. It either lacks the required elements or presents information that contradicts the prompt.

\textit{2 (Has significant discrepancies)}: There are major mismatches between the task prompt and the model's response. Key elements may be missing, or relationships and attributes are significantly misrepresented.

\textit{3 (Has several minor discrepancies)}: The response aligns with the prompt in a general sense but has noticeable errors or omissions in key details. While the main elements are present, finer aspects are not accurately rendered.

\textit{4 (Has a few minor discrepancies)}: The response is mostly aligned with the prompt, with only a few small and less critical details being inaccurate or missing. The response matches the description well but not perfectly.

\textit{5 (Matches exactly)}: The model's response perfectly matches the task prompt. All elements, relationships, attributes, and details align seamlessly with the description, with no noticeable discrepancies.

Now, return a score between 1 and 5 based on your memory of \texttt{\{self.model\_name\}}'s capabilities and limitations. Return a single number only, nothing else.
\end{tcolorbox}

\begin{tcolorbox}[colback=gray!10, colframe=gray!60, coltitle=black,
title=Predict Score - \tm, fonttitle=\bfseries, breakable]
Your current memory (knowledge) of the tool \texttt{\{model\_name\}} is: ``\texttt{\{current\_memory\}}'' \\
Based on your memory of model \texttt{\{model\_name\}}, predict the score for the following prompt: ``\texttt{\{prompt\}}''. \\

The scoring metric used here is a 5-point Likert scale: \\
\textit{1 (Does not match at all)}: The generated image or video completely fails to align with the text prompt. It either lacks the required elements or depicts a scenario that contradicts the prompt. \\
\textit{2 (Has significant discrepancies)}: There are major mismatches between the text prompt and the visual output. Key elements may be missing, or relationships and attributes are significantly misrepresented. \\
\textit{3 (Has several minor discrepancies)}: The output aligns with the prompt in a general sense but has noticeable errors or omissions in key details. While the main elements are present, finer aspects are not accurately rendered. \\
\textit{4 (Has a few minor discrepancies)}: The output is mostly aligned with the prompt, with only a few small and less critical details being inaccurate or missing. The visual matches the description well but not perfectly. \\
\textit{5 (Matches exactly)}: The generated image or video perfectly matches the text prompt. All elements, relationships, attributes, and details align seamlessly with the description, with no noticeable discrepancies. \\

Now return a score between 1 and 5 for prompt \texttt{\{prompt\}}, based on your memory of the tool \texttt{\{model\_name\}}. Return a single number only, nothing else. \\
\end{tcolorbox}

\subsection{Text Generation Tools}

\begin{tcolorbox}[colback=gray!10, colframe=gray!60, coltitle=black,
title=Prompt for Memory Refinement, fonttitle=\bfseries, breakable]
You are an assistant responsible for maintaining an incremental memory of the tool's capabilities, limitations, and use cases. \\
You will receive a current memory and new feedback. Your job is to update the memory by integrating the new feedback without losing previously stored information. \\

\textbf{Rules:} \\
1. Start with the current memory. \\
2. Consider the new prompt and feedback, but do not simply repeat the feedback, think deep and get general insights. \\
3. Incrementally refine and update the memory to reflect any new insights. \\
4. Do not remove previously known capabilities unless the new feedback explicitly contradicts them. \\
5. If no new insights are provided, leave the memory unchanged. \\

\textbf{Instructions:} \\
Follow this procedure strictly: \\
- Review the current memory for model \texttt{\{tool\_name\}}: \\
\quad ``\texttt{\{current\_memory\}}'' \\
- Input task prompt to the tool was: \\
\quad ``\texttt{\{task\_prompt\}}'' \\
- Tool's answer was: \\
\quad ``\texttt{\{response\}}'' \\
- Score rubric was: \\
\quad ``\texttt{\{score\_rubric\}}'' \\
- The score for tool's response was: \\
\quad ``\texttt{\{score\}}'' \\
- Feedback for this answer received: \\
\quad ``\texttt{\{feedback\}}'' \\

Integrate the key insights of the tool's limitations and capabilities from the feedback into the existing memory. Use complete sentences starting with phrases like \texttt{good at}, \texttt{bad at}, \texttt{proficient at}, or \texttt{poor at} to describe your findings. \\
- If the feedback suggests new limitations, capabilities, or clarifications, add them. \\
- If the feedback suggests that something previously stated is incorrect, correct it. \\
- If the feedback suggests the tool struggles with a certain aspect (like counting multiple objects), note it. \\
- If no contradiction or new insight is provided, keep the memory as is. \\
- Think deeply and determine what the feedback actually means for the model's capability, rather than staying on the surface of a specific example. \\

In your final answer, output \textbf{ONLY} the updated overall memory of the tool. Do not include extra explanations. \\
\end{tcolorbox}

\begin{tcolorbox}[colback=gray!10, colframe=gray!60, coltitle=black,
title=Predict Score - \gr, fonttitle=\bfseries, breakable]
Your current memory (knowledge) of the tool \texttt{\{model\_name\}} is: ``A large language model'' \\
Based on your memory of model \texttt{\{model\_name\}}, predict the score this model will get for answering the following task prompt: ``\texttt{\{prompt\}}''. \\

The scoring metric used here is a 5-point Likert scale: \\

\texttt{\{rubric\}} \\

Now return a score between 1 and 5, based on your memory of \texttt{\{model\_name\}}'s capabilities and limitations. \\
Return a single number only, nothing else.
\end{tcolorbox}

\begin{tcolorbox}[colback=gray!10, colframe=gray!60, coltitle=black,
title=Predict Score - \fs, fonttitle=\bfseries, breakable]
Your current memory (knowledge) of the tool \texttt{\{model\_name\}} is: A large language model. \\
Here are some examples of the model's performance on some tasks: \\

Task Prompt: \texttt{\{task\_prompt\}} \\
Rubric: \texttt{\{rubric\}} \\
Model's Score: \texttt{\{score\}} \\

Task Prompt: \texttt{\{task\_prompt\}} \\
Rubric: \texttt{\{rubric\}} \\
Model's Score: \texttt{\{score\}} \\

Task Prompt: \texttt{\{task\_prompt\}} \\
Rubric: \texttt{\{rubric\}} \\
Model's Score: \texttt{\{score\}} \\
...

Based on those samples and your memory of \texttt{\{model\_name\}}, predict the score this model will get for answering the following task prompt: ``\texttt{\{task\_prompt\}}''. \\

The scoring metric used here is a 5-point Likert scale: \\
\texttt{\{rubric\}} \\

Now return a score between 1 and 5, based on your memory of \texttt{\{model\_name\}}'s capabilities and limitations. \\
Return a single number only, nothing else.
\end{tcolorbox}

\begin{tcolorbox}[colback=gray!10, colframe=gray!60, coltitle=black,
title=Predict Score - \tm, fonttitle=\bfseries, breakable]
Your current memory (knowledge) of the tool \texttt{\{model\_name\}} is: \texttt{\{current\_memory\}} \\
Based on your memory of model \texttt{\{model\_name\}}, predict the score this model will get for answering the following task prompt: ``\texttt{\{prompt\}}''. \\

The scoring metric used here is a 5-point Likert scale: \\

\texttt{\{rubric\}} \\

Now return a score between 1 and 5, based on your memory of \texttt{\{model\_name\}}'s capabilities and limitations. \\
Return a single number only, nothing else.
\end{tcolorbox}

\begin{figure}[htbp]
    \centering
    \includegraphics[width=\linewidth]{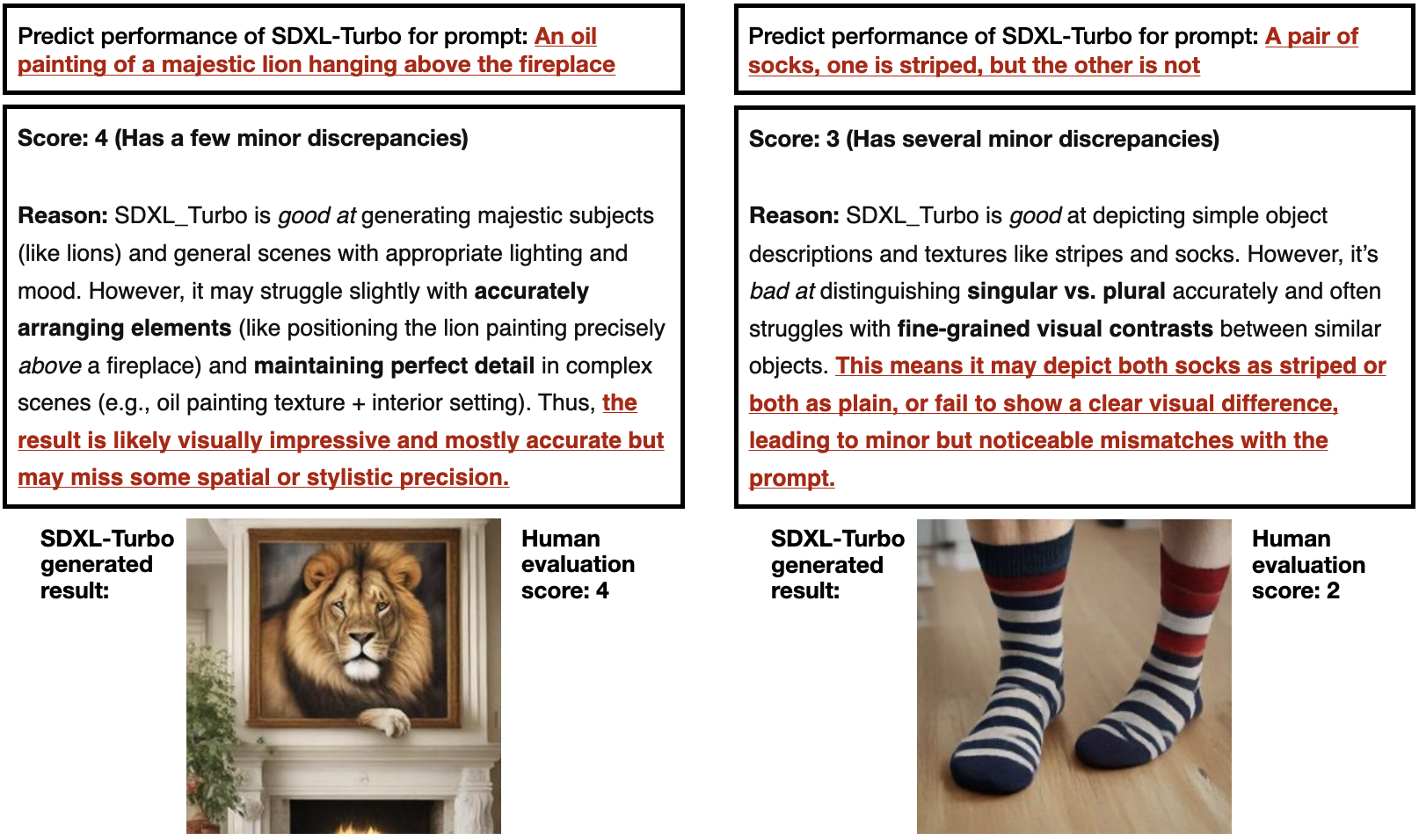}
    \caption{Qualitative examples for \tm. The left image illustrates the prediction for ``An oil painting of a majestic lion hanging above the fireplace,'' while the right example shows the prediction for ``A pair of socks, one is striped, but the other is not.''}
    \label{fig:case_study}
\end{figure}

\section{Case Study}
\label{app:case_study}

To better understand \tm's capability in performance prediction, we present two representative examples in Figure~\ref{fig:case_study}, where \tm is prompted to explicitly forecast the quality and fidelity of the output images generated by the SDXL-Turbo tool.

\paragraph{Example 1: Majestic Lion Above the Fireplace.}
For the prompt ``\emph{An oil painting of a majestic lion hanging above the fireplace},'' \tm assigns a predicted score of 4 (minor discrepancies). In practice, SDXL-Turbo successfully produces a visually appealing scene with a detailed lion portrait. However, our system anticipates that fine spatial alignment may not be perfectly captured (e.g., the fireplace might be cropped or misaligned). Indeed, the actual generated image exhibits a slightly off positioning of the lion, confirming \tm's prediction. Notably, the final score aligns closely with human evaluations, underlining \tm's proficiency in assessing model outputs.

\paragraph{Example 2: Pair of Socks with Different Patterns.}
For the prompt ``\emph{A pair of socks, one is striped, but the other is not},'' \tm predicts a score of 3 (several minor discrepancies). While SDXL-Turbo generally handles simple object prompts and textures adequately, \tm notes the model’s recurring difficulty in distinguishing subtle visual contrasts between highly similar objects. In the actual output, both socks often appear identical (both striped), matching \tm’s forecast. However, human evaluators deemed these inaccuracies more severe, giving a lower score of 2. This discrepancy suggests that although \tm detects the correct type of error, it occasionally assigns higher tolerance to such mismatches than human annotators.

\paragraph{Discussion.}
These examples highlight both the strengths and limitations of \tm. On one hand, \tm demonstrates strong alignment with human judgments when identifying whether the main elements of a prompt are preserved or substantially distorted. On the other hand, its scoring mechanism sometimes diverges from that of human evaluators, underscoring the subjective nature of visual quality assessments. 
\section{Compute Resource and Reproducibility Details}
\label{app:computation}

The core computation in our experiments is conducted via OpenAI’s API, GPT-4o-mini serves as the backbone language model for \tm, and text-embedding-ada-002 for RAG tasks. As our framework relies on API calls, it does not require any local GPU hardware for inferences.

All other models involved---such as Claude-3, DALL·E 3, Midjourney, SDXL, and others---are part of the precomputed outputs provided by the \gb and \bb datasets. We do not rerun these tools; we only analyze and build memory around their existing outputs.

The only locally executed model in our pipeline is the clip-flant5-xl model used to calculate VQA Score, which evaluates image-description alignment. This model is run on an NVIDIA A100 GPU, and is highly efficient—each VQA score can be computed in about 0.2 seconds. As such, the compute requirement is minimal, and large batches can be processed in real time.

Overall, our pipeline is lightweight and fully reproducible. All heavy model inference is handled externally via OpenAI APIs, and the only local requirement is for lightweight VQA scoring, making \tm practical to deploy and evaluate with modest resources.
\section{Broader Impact and Limitation}
\label{app:impact_limitation}

This paper investigates a new capability for agents powered by Large Language Models (LLMs): building and utilizing structured memory representations of tool capabilities to improve tool selection and task-solving performance. By enabling agents to systematically record and retrieve performance feedback from past tool interactions, our proposed \tm framework pushes toward more autonomous and adaptive decision-making systems.

The broader impact of this line of work lies in its potential to make complex multi-tool reasoning more robust, transparent, and efficient. Applications that involve numerous generative or decision-support tools—such as multimodal assistants, educational platforms, or creative design systems—could benefit from better memory-guided tool selection, reducing trial-and-error usage and model waste. \tm may also lower the barrier for non-experts to leverage advanced model capabilities, as the system autonomously adapts based on prior observations.

However, several limitations and risks must be acknowledged. First, the memory quality heavily depends on the accuracy and reliability of the feedback used to construct it. In this work, we rely on GPT-4-generated scores and rubrics, which, although effective, may introduce biases or inconsistencies not present in human-grounded evaluations. Moreover, the framework assumes tool capabilities remain relatively stable over time. In real-world deployments, frequent model updates or usage context shifts may render historical memory entries stale or misleading.

Another concern is over-reliance on memorized preferences, which may lead to suboptimal or unfair tool choices in edge cases not well-represented in the agent’s prior experiences. Without proper monitoring, such systems might reinforce early biases or miss the opportunity to explore potentially better alternatives.

Lastly, our experiments are conducted in controlled benchmark environments with known tool outputs and synthetic feedback. Generalizing \tm to open-ended tasks, dynamic tool ecosystems, or safety-critical applications requires additional safeguards and validation, which we leave for future work.

While promising, \tm represents an early step toward memory-augmented tool reasoning, and further exploration is needed to ensure both its robustness and its alignment with human-centered values in broader deployment contexts.

\end{document}